# Low-Resource Fine-Tuning for Multi-Task Structured Information Extraction with a 1B Instruction-Tuned Model Significance Testing and Output Quality Evaluation

Yu-Cheng Chih *

*International Intercollegiate Ph.D. Program. National Tsing Hua University, Taiwan*

*NLP Algorithm Engineer*

*Former Engineer, Accton Technology Corporation*

Email：s111003816@m111.nthu.edu.tw

Yong-Hao Hou

*Department of Computer Science, University of Taipei, Taiwan*

*AI Algorithm Engineer*

Email：g11016011@go.utaipei.edu.tw

## Abstract

Deploying large language models (LLMs) for structured data extraction in domains such as financial compliance reporting, legal document analytics, and multilingual knowledge base construction is often impractical for smaller teams due to the high cost of running large architectures and the difficulty of preparing large, high-quality datasets. Most recent instruction-tuning studies focus on 7B-parameter or larger models, leaving limited evidence on whether much smaller models can work reliably under low-resource, multi-task conditions. This work presents ETLCH, a 1B-parameter LLaMA-based model fine-tuned with low-rank adaptation on only 100–1000 samples per task for JSON extraction, knowledge graph extraction, and named entity recognition. Despite its small scale, ETLCH outperforms strong baselines across most evaluation metrics, with substantial gains observed even at the lowest data scale. These findings demonstrate that well-tuned small models can deliver stable and accurate structured outputs at a fraction of the computational cost, enabling cost-effective and reliable information extraction pipelines in resource-constrained environments.

# 1. Introduction

Since the advent of large language models (LLMs), generative artificial intelligence (Generative AI) has profoundly reshaped industrial practices in enterprise data extraction, regulatory compliance automation, multilingual knowledge management, and real-time decision support systems. Despite these advances, deploying most models into real-world applications continues to face two critical bottlenecks: (1) the difficulty of acquiring large volumes of high-quality annotated data, and (2) the prohibitive computational resources required for large-scale architectures [3]. These limitations are particularly acute in data-rich yet unstructured environments—such as financial auditing records, medical case reports, legal proceedings, and cross-lingual corporate arc-hives—where converting raw data into machine-usable structured formats for fine-tuning often relies on labor-intensive processes, including manual field parsing and entity extraction. Such workflows create substantial barriers to the automation of knowledge transformation and large-scale ETL (Extract–Transform–Load) pipelines.

Current research has predominantly advanced along two main trajectories. The first focuses on pure prompt engineering, leveraging the natural language understanding capabilities of LLMs to guide data extraction and automatically generate standardized outputs such as JSON. While this approach can achieve relatively high accuracy with commercial-scale proprietary models such as ChatGPT, smaller on-premise LLMs—which are preferred for privacy compliance, edge deployment, and cost control—remain constrained by architectural limitations, resulting in unstable instruction adherence, inconsistent structured outputs, and limited batch-processing scalability [4]. The second trajectory involves instruction fine-tuning, where multi-task, multi-instruction annotations are used to train models for specific structured data extraction workflows. Studies published in top-tier venues have demonstrated that this method improves comprehension of target instructions and enhances the consistency and controllability of structured outputs. However, such benefits typically depend on large models (e.g., LLaMA-7B or above) trained with tens of thousands of samples on multiple A100 GPUs, making the approach financially and technically inaccessible to small and medium-sized enterprises (SMEs) and research teams without dedicated AI infrastructure [3], [4].

To date, systematic experiments examining the combination of small-scale LLMs (e.g., 1B parameters) with cost-efficient fine-tuning strategies—such as low-rank adaptation

(LoRA) at modest ranks—under low-resource, multi-task, industry-oriented instruction scenarios remain scarce [1], [3], [9]. To address this gap, this study builds on a 1B-parameter lightweight LLM, integrating low-rank LoRA fine-tuning into a multi-instruction, multi-task training framework. The proposed approach emphasizes data efficiency, domain adaptability, and operational stability, enabling automated structured information extraction in domains such as financial compliance reporting, medical record digitization, legal document analytics, and multilingual enterprise knowledge integration. By demonstrating that a compact 1B model can match or exceed the extraction accuracy of much larger LLMs in these scenarios, this work establishes a practical and cost-effective paradigm for the democratization of intelligent ETL systems, lowering the entry barrier for AI adoption in SMEs and domain-specific data environments.

## 2. Related Work

The rapid development of generative AI technologies has unlocked substantial po-tential for application innovation. However, realizing concrete functionality typically presupposes access to large-scale, high-quality annotated datasets, making the design of data features and the conversion into fine-tuning-ready formats a persistent challenge [3]. Even when rich data sources are available, not all organizations possess the capacity to transform them into effective fine-tuning formats. Consequently, the ability to perform structured element extraction has emerged as a central concern for both research and industry.

Existing approaches can be broadly categorized into two directions. The first relies on prompt-based guidance, wherein the model is instructed to extract data and output it in structured formats such as JSON. While this method exhibits relatively stable performance in commercial-scale models (e.g., ChatGPT), it suffers from instability in locally deployed models due to inherent limitations in their weight architectures. This instability manifests as inconsistent results for identical prompts across different texts, rendering such methods unreliable for large-scale ETL (Extract–Transform–Load) applications [4]. The second category focuses on instruction fine-tuning, in which model parameters are optimized for specific instruction formats, thereby significantly improving proficiency in task-specific instructions and the accuracy of structured data extraction [3], [4]. Frameworks that combine multi-task, multi-instruction training with structured output annotation have been shown to enhance consistency, controllability, and diversity in generated outputs [3], [9].

Notably, state-of-the-art studies typically adopt large models—such as LLaMA-7B or higher (e.g., 13B, 33B, 65B), LLaMA-2, or GPT-3—trained on tens of thousands of examples using high-end computational resources (e.g., 8×A100 40GB GPUs). For example, JsonTuning utilized multi-task datasets such as Flan 2022 (~50k samples) and InstructUIE (~10k samples), all standardized to JSON outputs, training a LLaMA-7B model on multiple A100 GPUs [3]. Similarly, InstructIE employed 14,579 diverse instruction–output pairs, fine-tuning LLaMA-7B with LoRA (rank up to 16) on an 8×A100 setup to enable efficient instruction-based structured extraction [4].

Other notable work includes Learning to Extract Structured Entities Using Language Models (EMNLP 2024), which integrates multiple NER, relation extraction, and event annotation datasets—ranging from several thousand to tens of thousands of sam-ples—and conducts experiments on T5-Large, LLaMA-7B, or equivalent transformer architectures using multi-GPU A100/V100 environments, with all annotations standardized to JSON or schematic formats [9]. In Nature Communications (2024), Dagdelenet al.. addressed scientific text extraction using the GPT-3 API and LLaMA-2 LoRA (7B/13B) on a combination of 2×A100 80GB GPUs and cloud environments, with datasets of 500–3,000 structured annotations validated across multiple output formats [1].

## Limitations of 7B/8B-scale LLMs for Resource-Constrained Users

Despite the extensive empirical evidence available for 7B/8B-scale models in the open-source community, these approaches remain largely inaccessible to users with limited resources due to several factors:

- Large model size – A 7B model's full weights alone require approximately 14–24 GB of VRAM, with even higher requirements for 8B models, exceeding the capacity of typical consumer GPUs (e.g., RTX 3060/4060) [2], [3].

- High computational requirements – Most state-of-the-art studies assume access to multiple A100/V100 data center–class GPUs or equivalent large-scale cloud infrastructure. Without such resources, training becomes prohibitively slow or infeasible [1], [3], [4].

- Extensive data requirements – Existing designs often assume tens of thousands of training samples, making them impractical for researchers or enterprises lacking large annotated corpora and the resources to build them [3], [4], [7].

- Challenges in deployment and maintenance – Even with successful fine-tuning, the end-to-end deployment of such large models (e.g., integrating into ETL pipelines or enterprise APIs) incurs significantly higher costs compared to smaller-scale LLMs (e.g., 1B parameters), limiting their adoption in practice [7].

A review of the literature further reveals that most work has focused on English corpora and publicly available large models, leaving a gap in systematic and large-scale evaluations of small-scale LLMs (e.g., 1B parameters) combined with low-cost fine-tuning strategies (e.g., low-rank LoRA) in low-resource, multi-task instruction scenarios [1], [3], [5], [9], [10]. For resource-constrained or language-specific applications, there remains a lack of empirical evidence and practical technical guidelines.

To address these gaps, this study investigates the integration of low-rank LoRA fine-tuning with a 1B-parameter lightweight LLM under extremely constrained training resources, incorporating a multi-instruction, multi-task design. The aim is to evaluate the feasibility and identify best practices for automated structured information extraction, thereby extending and complementing the current body of literature.

## 3. Method and Data

### 3.1. Parameter Design and Training Environment

All experiments in this study were conducted on a single consumer-grade GPU, with the setup designed to ensure smooth execution under such hardware conditions. At peak training load, a single NVIDIA RTX 3090 was sufficient to support the entire computational process. The base model used was Llama-3.2-1B-Instruct (Hugging Face model name: meta-llama/Llama-3.2-1B-Instruct).

This model was chosen for its relatively small parameter size, which allows deployment in resource-constrained environments. While its knowledge precision is inherently limited compared to larger LLMs, it is capable of understanding natural language intent and maintaining coherent dialogue without producing irrelevant or nonsensical responses. However, its factual accuracy is less reliable, necessitating task-specific fine-tuning.

To align with the study's low-resource design, LoRA fine-tuning was applied without quantization, given the model's already modest parameter count. The LoRA configuration adopted a rank of 32 and α = 64. To mitigate overfitting in low-data regimes, the effective batch size was set to 2, thereby introducing higher noise levels. A dropout rate of 0.4 was further applied to encourage the model to learn relevant data patterns even under noisy conditions.

To avoid significant divergence from the pretrained weight distribution, parameter updates were kept minimal, with a **learning rate of $1 \times 10^{-7}$** and a **maximum gradient norm of 0.1**. For comparability across experiments, all configurations used **100 epochs**. It should be noted that while the number of epochs was fixed, differences in dataset sizes inherently led to varying numbers of parameter updates; thus, "100 epochs" here denotes the number of full passes over each dataset, not an identical number of update steps.

### 3.2. Data

The instruction tuning dataset in this study uses long-form knowledge-based texts generated by the ChatGPT API (gpt-4o-mini model) as *context*. Each text is relatively long, with some reaching up to about 1,500 tokens in length, and covers diverse and randomly selected topics such as quantum mechanics, the golden ratio in mathematics, and artificial intelligence. This design ensures diversity in both contextual depth and knowledge domains within the training samples, which helps the model learn the ability to perform structured information extraction across different topics.

### 3.3. Planned Demonstration Tasks and Selection Rationale

The planned demonstration extraction tasks in this study are JSON extraction, Knowledge Graph Extraction (KGE), and Named Entity Recognition (NER). The rationale for selecting these three tasks is as follows:

### 3.4. A reasoning chain from basic to advanced

Reasoning: If the model cannot accurately identify key entities in the text (NER), then any subsequent derivation of semantic structures cannot be performed correctly. Therefore, NER must be the first step in the reasoning chain. Once entities are correctly identified, the next reasoning step is to establish the semantic relationships among them (KGE), which enables the model to perform multi-hop reasoning and contextual relationship analysis. Finally, these reasoning-organized entities and relationships need to be converted into structured data (JSON extraction) that can be directly processed by a system for practical applications. This design forms a complete reasoning chain of "entity recognition → relationship construction → structured output."

### 3.5. Mutual verification and reinforcement between tasks

Reasoning: When the three tasks are trained simultaneously, the model's output for one task is constrained by the results of the others. For example, if the entities output by NER do not match the subject entities in KGE, the model will self-correct during training due to loss penalties. The format constraints of JSON extraction also force the model to maintain logical consistency in the final output. This multi-layer cross-task reasoning verification mechanism is, in reasoning terms, able to reduce the risk of error accumulation and amplification within the reasoning chain.

### 3.6. Alignment with real-world industrial reasoning needs

Fact: In domains such as financial compliance checks, legal analysis, medical record digitization, and knowledge retrieval, the data processing workflow often follows the steps of "extract key entities → derive relationships → produce actionable data."

Reasoning: Since the sequence of these three tasks matches the logical reasoning chain in industry, it can be reasonably expected that jointly training them will make the model better adapted to real-world operational requirements.

### 3.7. Reasoning efficiency under resource-constrained conditions

Reasoning: Compared to training separate models for each task, integrating the three tasks into a single model can avoid information loss and delays caused by inter-model data transfer during the reasoning stage, while improving overall computational efficiency. This design, in reasoning terms, is particularly suitable for small and medium-sized enterprises and government agencies with limited computational resources, as it can reduce costs while retaining cross-task applicability.

## 4. Labeling the data

1 **Extraction to JSON:**

    1.1 Each annotation generated by ChatGPT consists of three clearly defined components:

- 1.1.1 A long-form, semantically dense article covering diverse and randomly selected knowledge topics (serving as the context, i.e., the generated text described in the Data section of this study).
- 1.2 An Instruction, generated based on the context in item 1, that explicitly specifies the information to be extracted and defines the required JSON output schema, including strict constraints such as requiring all values to be lists without nested structures.
- 1.3 A gold-standard JSON output, generated based on both the context in item 1 and the In-struction in item 2, containing only the extracted information and strictly adhering to the schema described in the Instruction.
- 1.4 Note: The semantically dense article in item (1) serves as the sole basis for item (2); item (3) is based on both item (1) and item (2).

## 2 Knowledge Graph (KGE): KGE data annotation instruction

The original training instructions are in Chinese, as the model was fine-tuned primarily for Chinese-language structured information extraction. An English translation is provided for clarity.

Original Chinese instruction:

「請幫我為下面文章建構知識圖譜。請以這種方式輸出：「【一、三元組格式（主詞－關係－受詞）】植物基飲食－可以降低－心臟病風險植物基飲食－可以降低－糖尿病風險植物基飲食－可以降低－癌症風險植物基飲食－有助於攝取－纖維素植物基飲食－有助於攝取－抗氧化物質健康生活－包含－飲食健康生活－包含－運動健康生活－包含－心理健康運動－能夠幫助－控制體重運動－能夠幫助－增加肌肉健康運動－能夠幫助－增加骨骼健康運動－能夠釋放－內啡肽運動－有助於改善－心理健康冥想－能夠幫助－減少壓力深呼吸－有助於－壓力減緩親近自然－能夠提升－生活品質」+context

English translation (for illustration only):

Please construct a knowledge graph for the following article. Output in the following format:

【1. Triple format (Subject–Relation–Object)】

Plant-based diet – can reduce – risk of heart disease

Plant-based diet – can reduce – risk of diabetes

Plant-based diet – can reduce – risk of cancer

Plant-based diet – helps intake – dietary fiber

Plant-based diet – helps intake – antioxidants

Healthy lifestyle – includes – diet

Healthy lifestyle – includes – exercise

Healthy lifestyle – includes – mental health

Exercise – can help – control weight

Exercise – can help – increase muscle

Exercise – can help – improve bone health

Exercise – can release – endorphins

Exercise – can improve – mental health

Meditation – can help – reduce stress

Deep breathing – helps – stress relief

Being close to nature – can improve – quality of life

+context (original article text follows)

## 3 Named Entity Recognition (NER) — Annotation Instruction

The original training instructions are in Chinese, as the model was fine-tuned primarily for Chinese-language structured information extraction. An English translation is provided for clarity.

Original Chinese instruction:

*請為下文執行 NER 任務。請輸出成 JSON 且 value 必須為 list 格式，而其中不得再有巢狀結構。*

*+context+先前階段由 ChatGPT 所產生的標註*

English translation (for illustration only):

Please perform a named entity recognition (NER) task for the following text. Output the re-sults in JSON format, ensuring that every value is a list and that no lists contain nested structures.

+context + the annotation labels produced by ChatGPT in the preceding stage (output)

## 5. Performance Evaluation

The model trained following the above methodology is referred to as ETLCH, while other open-source models are used in their original released versions.

ETLCH-XXX denotes the variant trained on XXX samples per task; across the three tasks, the combined train+validation size is 3×XXX.

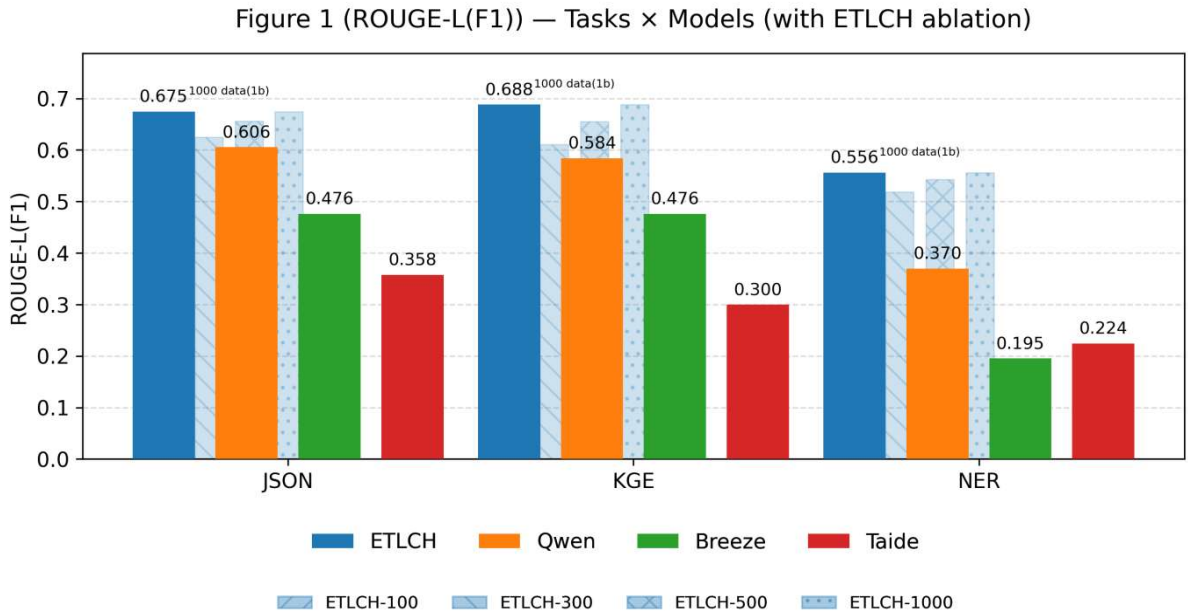

Figure 1. ROUGE scores of different models on the test sets of the three aforementioned tasks

From Figure 1, Llama-3.2-1B-Instruct—referred to in this paper as ETLCH after fine-tuning—achieves the highest ROUGE-L (F1) scores across all three tasks when trained on only 100 samples for 100 epochs. For the JSON and KGE tasks, ETLCH delivers substantial gains over the state-of-the-art open-source baseline, with statistical tests confirming significance (KGE – ROUGE-L (F1): $p = 3.97 \times 10^{-124}$; JSON – ROUGE-L (F1): $p = 3.09 \times 10^{-30}$; both $< 0.05$). In the NER task, although ETLCH attains the highest ROUGE-L (F1) score, the difference from the baseline model Qwen2.5 is not statistically significant ($p = 0.053$), indicating comparable performance under this setting. Statistical significance was assessed at the 0.05 level.

When the dataset size increases beyond 100 samples (i.e., 300, 500, or 1000 samples), ETLCH maintains higher scores than general-purpose pretrained models in JSON and KGE, with these advantages remaining sustained and stable, while its NER performance advantage also becomes more pronounced.

Among the evaluation metrics, ROUGE-L is particularly noteworthy. It measures the proportion of the Longest Common Subsequence (LCS) between the model-generated text and the reference text, incorporating both recall and precision, and reports a com-bined F1 score.

Compared with ROUGE-1 (unigram matching) and ROUGE-2 (bigram matching), ROUGE-L places greater emphasis on sentence structure and word order consistency, thereby reflecting how well the generated text preserves the context and structure of the source. Under the low-resource setting of 100 training samples, ETLCH attains the highest ROUGE-L scores across all three tasks, indicating not only strong lexical coverage but also a high degree of syntactic alignment with the reference out-puts—demonstrating precise handling of both context and word order.

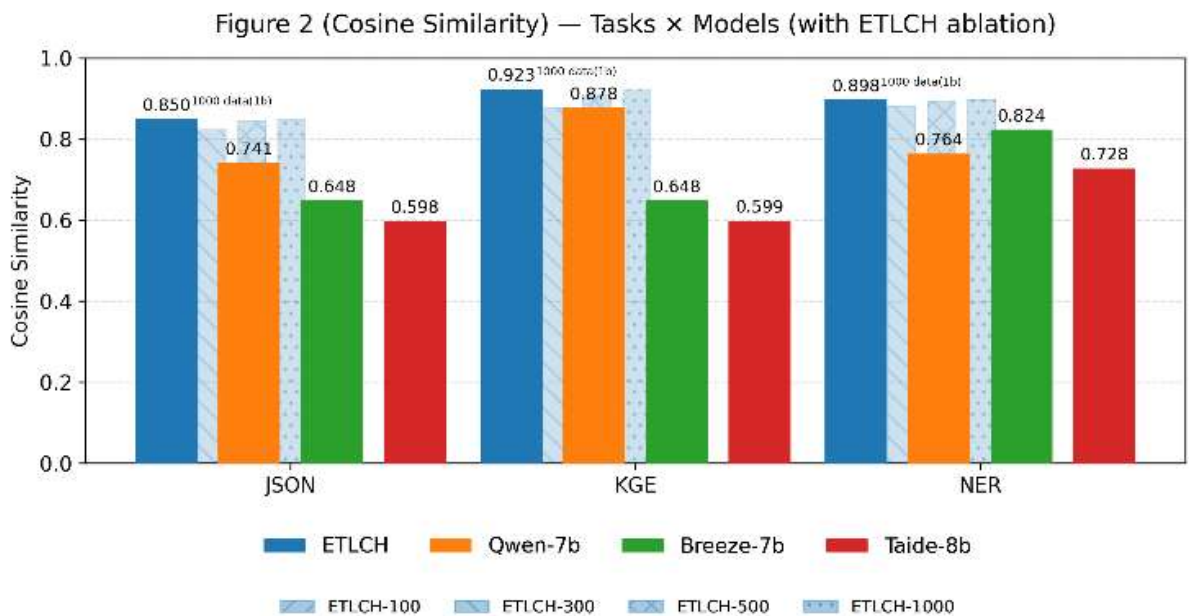

Figure 2. Cosine similarity between model outputs and ground truth texts on the test sets of the three aforementioned tasks

From Figure 2 (Cosine Similarity), statistical tests show: JSON – Cosine Similarity: $p = 5.35\times10^{-6}$; KGE – Cosine Similarity: $p = 3.34\times10^{-122}$; NER – Cosine Similarity: $p = 0.053$, not significant.

Statistical tests show that for the JSON task, $p = 5.35\times10^{-6}$, and for the KGE task, $p = 3.34\times10^{-122}$—both below the 0.05 threshold, indicating a significant advantage over Qwen. In contrast, for the NER task, $p = 0.053$, which does not meet the significance threshold, suggesting that the observed difference may be attributable to random variation, and the two models perform comparably.

In other words, ETLCH demonstrates a clear advantage in JSON and KGE tasks, whereas in the NER task, its performance is essentially on par with Qwen2.5-7B.

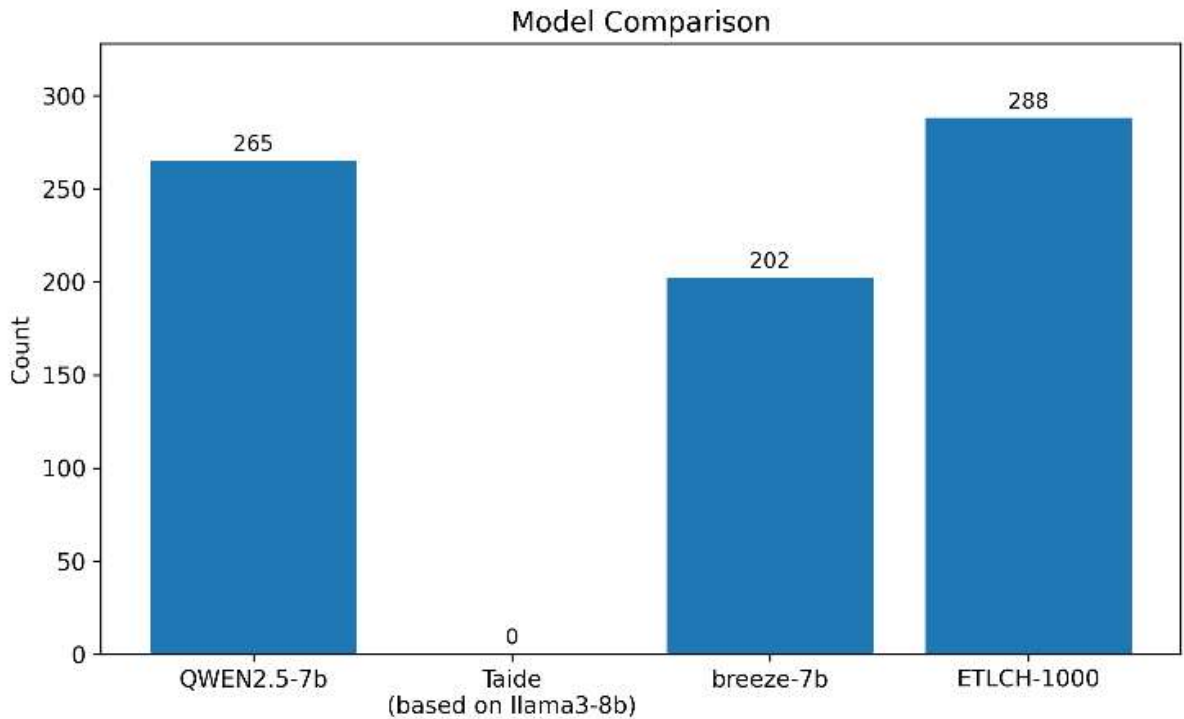

Figure 3. Number of JSON strings per model successfully parsed (json.loads)

It is striking that the Taiwan-localized large language model, fine-tuned from LLa-MA3-8B by the National Science and Technology Council, failed to produce a single correctly readable JSON string in the evaluation task. In sharp contrast, ETLCH—despite having only 1B parameters—outperformed the 8B-parameter Taide model, the state-of-the-art Qwen2.5-7B, and the closely ranked Breeze-7B after fine-tuning.

A two-proportion z-test against Taide-8B showed an exceptionally large difference ($p = 1.061 \times 10^{-158}$, $n = 300$). When compared with Qwen2.5-7B, the difference remained statistically significant ($p = 3.389 \times 10^{-4}$, $n = 300$), and against Breeze-7B, the gap was again extremely significant ($p = 1.687 \times 10^{-21}$, $n = 300$). These findings underscore the method's clear advantage in producing JSON with guaranteed readability, and more broadly, reveal the underappreciated potential of small-scale language models in structured output generation.

## 6. Data Efficiency Analysis

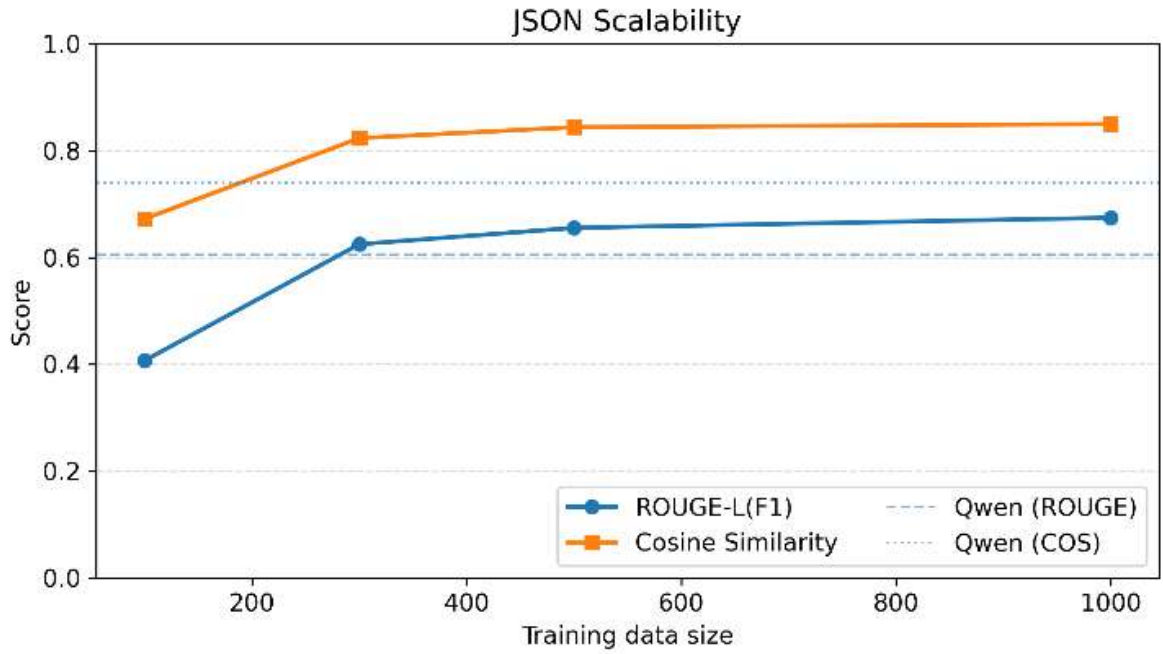

Figure 4. Effect of Data Volume on Improving Model Performance in the JSON Task

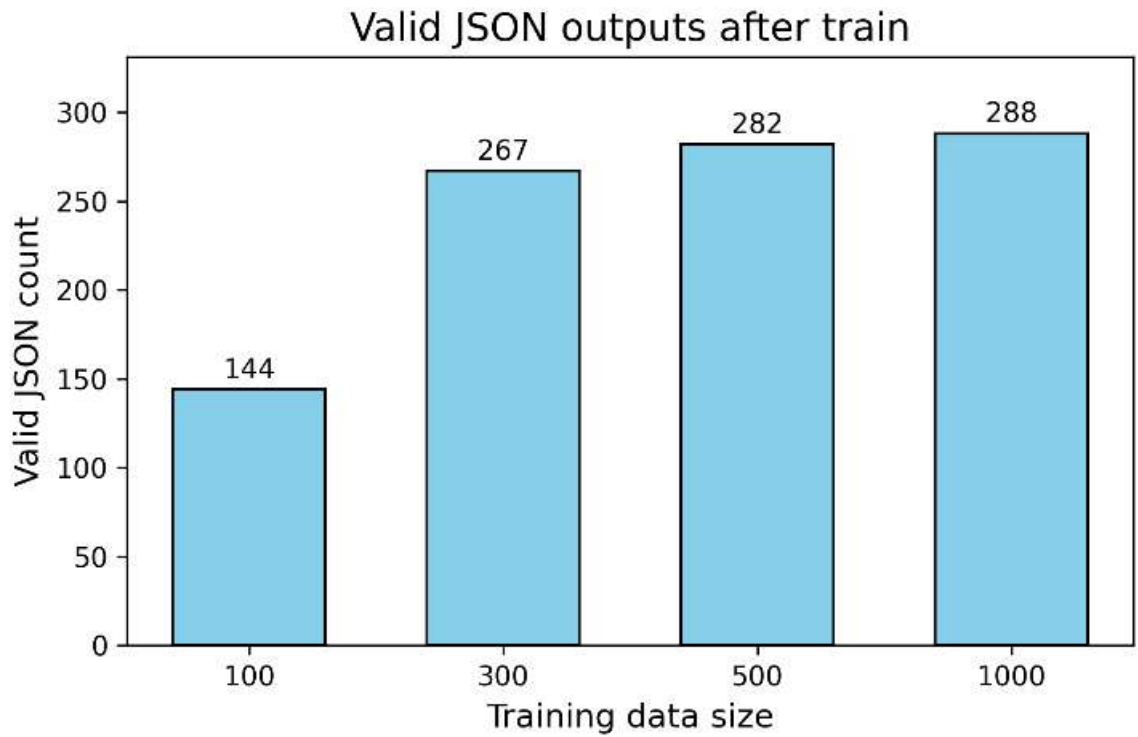

Figure 5. Valid JSON Outputs After Training

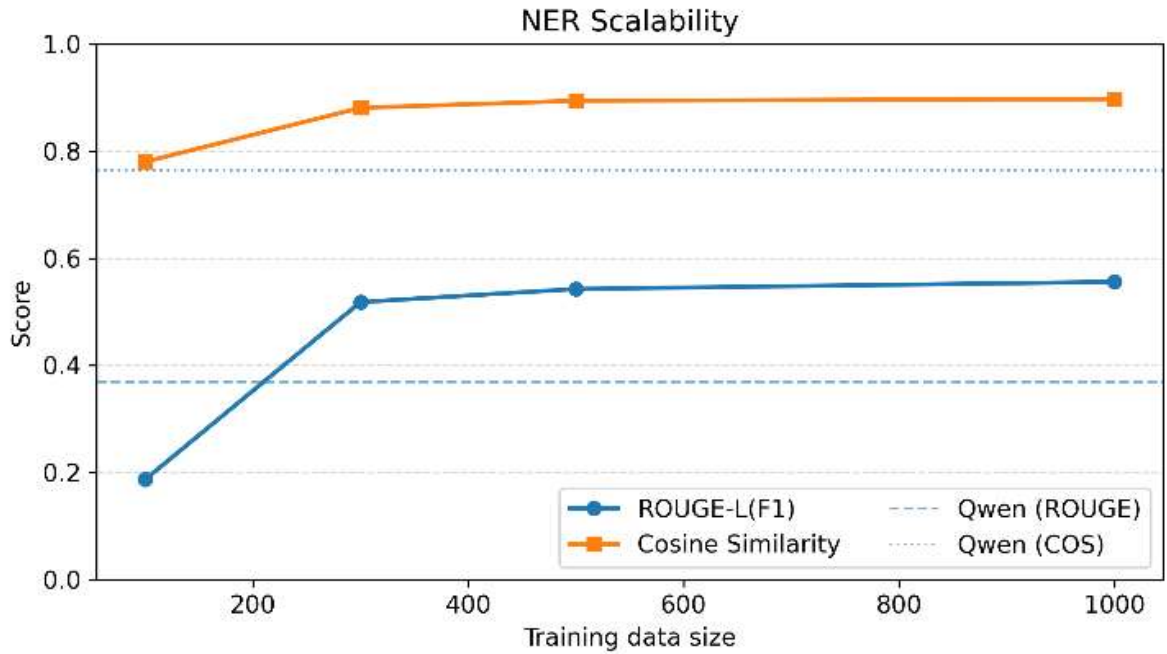

Figure 6. Effect of Data Volume on Improving Model Performance in the KGE Task

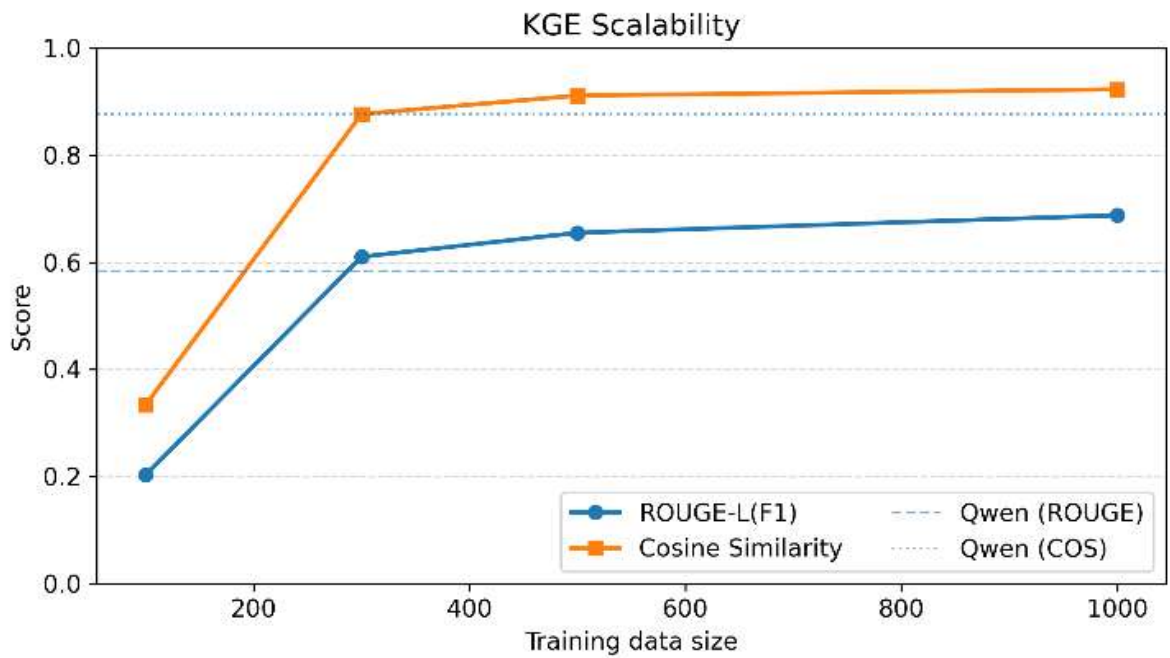

Figure 7. Effect of Data Volume on Improving Model Performance in the NER Task

As shown in Figures 4–7, ETLCH exhibits rapid performance gains across multiple tasks when the training data size increases from 100 to 300 samples, followed by a clear plateau between 300 and 1000 samples. For the JSON extraction, knowledge graph extraction (KGE), and named entity recognition (NER) tasks, both ROUGE-L (F1) and Cosine Similarity metrics improve substantially before 300 samples, after which the gains diminish markedly.

Compared to the strong Chinese-oriented instruction-tuned baseline Qwen2.5-7B[5], ETLCH matches or surpasses its scores on most tasks when the training data size reaches 300 samples or more, highlighting its data efficiency advantage. This pattern indicates that the model can quickly capture essential structural and semantic features under limited data conditions, and achieve notable improvements in low-resource scenarios.

Figure 5 further illustrates this trend: with 100 training samples, the model produces 144 valid JSON outputs successfully parsed by json.loads; when the data size increases to 300 samples, this number rises sharply to 267, and then increases only slightly to 282 and 288 at 500 and 1000 samples, respectively—demonstrating diminishing re-turns beyond 300 samples. In other words, ETLCH's sensitivity to data volume is concentrated in the low-resource regime, with limited per-formance gains once the data scale surpasses this threshold.

Overall, ETLCH demonstrates exceptional data efficiency in multi-task structured output generation. In practical applications, it can achieve near-plateau performance with a relatively modest amount of annotated data, underscoring its strong suitability for low-resource scenarios.

# 7. Conclusion

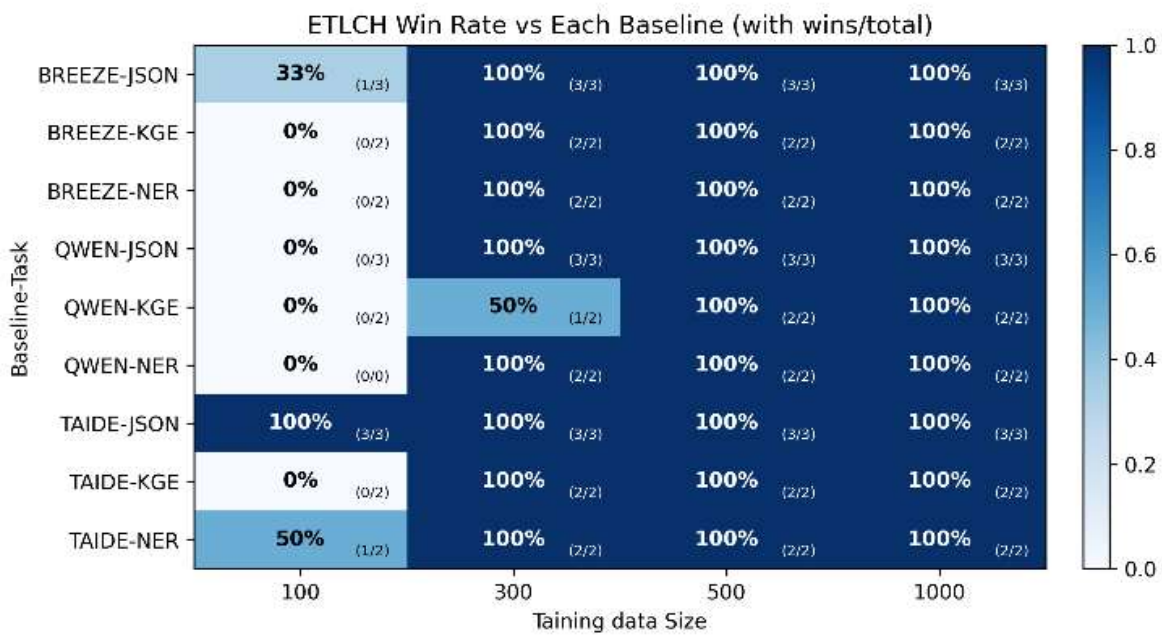

Figure 8. Winning rates of ETLCH on JSON extraction, Knowledge Graph Ex-traction (KGE), and Named Entity Recognition (NER) across different training data scales. JSON results are based on three metrics; KGE and NER use two metrics.

1 **Computation of Winning Rates (explicit "X out of Y wins")**
   1.1 For each combination of task × baseline model × training data size, we check each ap-plicable evaluation metric for that task to determine whether the proposed model outperforms the baseline (denoted as Win).
   1.2 Winning rate = Number of metrics with Win ÷ Total number of applicable metrics for that task.
      1.2.1 JSON extraction: Denominator = 3 (ROUGE-L F1, Cosine Similarity, JSON_Parse_OK — a binary indicator of whether the generated JSON output can be successfully parsed without errors). Example: 2/3 = 67%, 3/3 = 100%.
      1.2.2 KGE / NER: Denominator = 2 (ROUGE-L F1, Cosine Similarity). Example: 1/2 = 50%, 2/2 = 100%.
2 **Findings:**
   2.1 In most cases, ETLCH achieves outperformance rates above 70% against Breeze-7B, Qwen2.5-7B, and Taide.
   2.2 Even with only 100–300 training samples, the model maintains stable advantages in JSON extraction and NER across all data scales. While KGE shows greater variability, ETLCH remains competitive in most settings.
3 **Implications:**
   These findings, consistent with the statistical significance results in Section IV, demonstrate that small, well-tuned models can achieve structured information extraction performance comparable to or exceeding that of much larger models under severe resource constraints. This enables SMEs and specialized teams to attain high accuracy at lower computational and data costs. Moreover, the model can rapidly adapt to highly customized extraction requirements, supporting a wide range of structured output formats tailored to specific domains—such as financial compliance, medical records, and legal analytics—thereby extending applicability and flexibility across diverse real-world scenarios.
4 **Acknowledgment**
   This study focuses on exploring the performance of small-scale models, with all training conducted using the **LlamaFactory** framework [6],[8]. Additionally, AI tools were used for grammar correction, spelling revision, and sentence refinement.